\documentclass[10pt,twocolumn,letterpaper]{article}

\usepackage[pagenumbers]{cvpr} 

\usepackage{cuted}
\usepackage{booktabs}
\usepackage{float}
\usepackage{pgfplots}
\pgfplotsset{compat=1.18}

\definecolor{cvprblue}{rgb}{0.21,0.49,0.74}
\usepackage[pagebackref,breaklinks,colorlinks,allcolors=cvprblue]{hyperref}

\def\confName{CVPR}
\def\confYear{2027}

\title{TAO: Tracking-Aware Optimization for Camera Placement in Urban Scenes}

\author{Wei Geng\\
Renmin University of China\\
{\tt\small gengwei21@mails.ucas.ac.cn}
\and
Yongcai Wang\\
Renmin University of China\\
{\tt\small ycw@ruc.edu.cn}
\and
Jinlong E\\
Renmin University of China\\
{\tt\small ejinlong@ruc.edu.cn}
\and
Deying Li\\
Renmin University of China\\
{\tt\small deyingli@ruc.edu.cn}
\and
Yunjun Han\\
Institute of automation, Chinese academy of science\\
{\tt\small yunjun.han@ia.ac.cn}
}

\begin{document}
\maketitle
\begin{strip}
\centering
\includegraphics[width=0.95\textwidth]{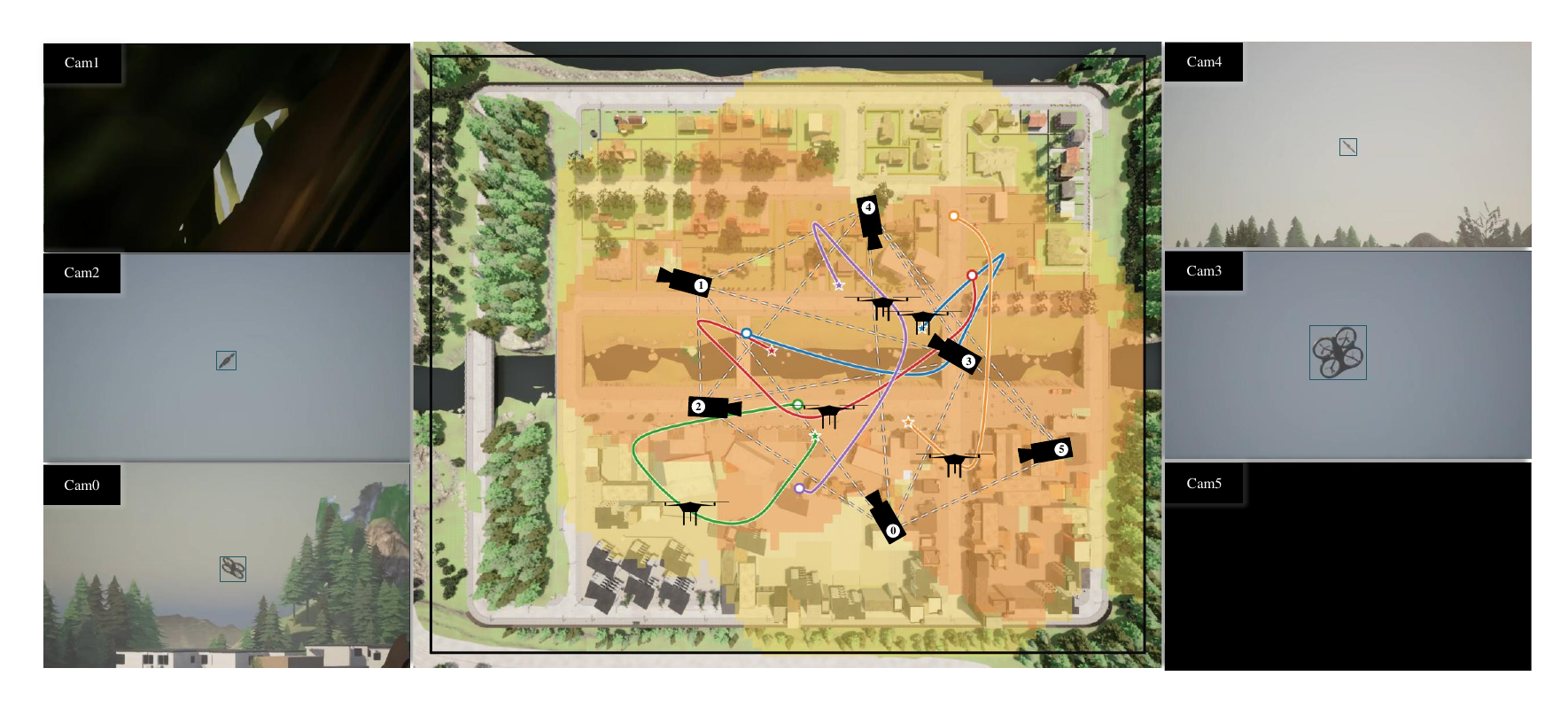}
\captionsetup{hypcap=false}
\captionof{figure}{\textbf{Overview of camera placement for tracking dynamic targets.} The top-down urban view shows a team of eight pan--tilt--zoom (PTZ) cameras tracking five autonomously moving UAV targets. Yellow and orange regions indicate the union of the cameras' visible areas, while orange regions indicate multiply covered areas where their fields of view (FoVs) overlap. The six images on the left and right are captured by six of these cameras, whose positions and viewing directions are marked in the top-down view.}
\captionsetup{hypcap=true}
\label{fig:overview}
\end{strip}

\begin{abstract}
\label{sec:abstract}
For camera placement, selecting informative viewpoints often enables camera systems to perform surveillance tasks more effectively, particularly for tracking dynamic targets in three-dimensional (3D) space. Continuous surveillance is generally more valuable than intermittent surveillance because it gives a surveillance system more time to anticipate target motion, coordinate camera handovers, and respond to events. We propose Tracking-Aware Optimization (TAO), a reinforcement learning (RL)-based deployment method for multiple pan--tilt--zoom (PTZ) cameras that promotes persistent tracking of moving targets. We further introduce per-target surveillance duration (PTSD), a unified metric that jointly measures the tracking ratio and continuous tracking duration, and derive a low-cost geometric surrogate, termed potential, through theoretical analysis. TAO derives its RL reward from the potential: robustness promotes overlapping fields of view (FoVs) for reliable cross-camera handovers, while coverage retention prevents robustness from sacrificing coverage too early. Together, these two components improve continuous tracking duration while preserving broad trackability. Finally, we train TAO by optimizing this surrogate on one CARLA map and evaluate the tracking performance of the resulting camera deployments on the other maps. Experiments across all eight urban maps validate the proposed method, metric, and surrogate: TAO improves PTSD, continuous tracking duration, and tracking ratio over competing deep-learning methods and even outperforms several search algorithms, while the ablation results confirm that both robustness and coverage retention contribute to its tracking performance.
\end{abstract}

\section{Introduction}
\label{sec:introduction}

Tracking multiple dynamic targets in three-dimensional (3D) space is crucial for urban management and public safety. Unlike targets in static scenes, both the motion trajectories and the number of dynamic targets can be unknown in advance. Pan--tilt--zoom (PTZ) cameras provide greater flexibility in viewing direction and focal length than fixed cameras, allowing limited camera resources to follow moving targets more efficiently. Their effectiveness, however, depends strongly on where the cameras are deployed.

Complex urban geometry introduces severe occlusions from buildings and other structures, which can drastically reduce a camera's usable field of view (FoV). Selecting appropriate deployment viewpoints is therefore essential for reliable tracking. A good viewpoint should not only provide an unobstructed view but also complement the FoVs of the other cameras to reduce blind regions. Consequently, multiple PTZ cameras are commonly treated as a coordinated team whose deployment is optimized jointly rather than as a collection of independent sensors. Figure~\ref{fig:overview} illustrates this coordinated camera-placement setting.

Camera-placement methods by H\"orster and Lienhart~\cite{HorsterLienhart2006} and Hanoun et al.~\cite{HanounEtAl2017} cast this task as a multi-camera coverage problem and optimize the union of the cameras' visible regions. As discussed by Yao et al.~\cite{YaoEtAl2008}, coverage is necessary but not sufficient for tracking dynamic targets. A coverage-oriented deployment can divide a target trajectory into many short tracklets separated by blind intervals. Such fragmented tracking is unfavorable for downstream tasks such as situational monitoring, trajectory prediction, and timely intervention (e.g., a single vehicle that repeatedly disappears and reappears may even be mistaken for multiple vehicles). Camera-placement optimization therefore requires an objective and evaluation metric that capture more than overall coverage. In addition to the fraction of a trajectory that is successfully tracked, they should measure how long each contiguous tracked segment persists.

Camera-placement policies must also adapt to different maps and target populations. Compared with handcrafted heuristics such as greedy search, deep learning (DL) offers a path toward learning deployment strategies that generalize across tasks and environments. Direct supervised learning is difficult, because the optimal camera layouts are rarely available as ground-truth labels. Reinforcement learning (RL) is well suited to this setting because it can optimize a deployment policy directly from task-level rewards. Prior work by Klinghoffer et al.~\cite{KlinghofferEtAl2023}, Chen et al.~\cite{ChenEtAl2021}, and Li and Li~\cite{LiLi2025} has likewise demonstrated RL-based optimization of camera configurations for task-dependent imaging, indoor depth observation, and surveillance coverage. However, these formulations primarily optimize task-specific perception accuracy or spatial coverage rather than the temporal continuity of unpredictable target trajectories. They also do not explicitly balance broad coverage against the overlapping FoVs needed for robust cross-camera handovers.

In this paper, we propose Tracking-Aware Optimization (TAO), an RL-based method for the joint deployment of multiple PTZ cameras to support persistent tracking of dynamic targets. We first introduce \emph{per-target surveillance duration (PTSD)}, a metric that jointly measures the tracking ratio of a trajectory and its continuous tracking duration. We then derive \emph{potential} as a low-cost geometric approximation of PTSD. The potential combines spatial coverage with deployment robustness, where robustness rewards overlapping FoVs that support continuous cross-camera handovers. TAO sequentially selects camera candidates using tracking-aware candidate-level and global observations and receives a dense reward equal to the incremental gain in a coverage-retained form of the potential. 
Together, robustness and coverage retention steer the policy toward placements with longer continuous tracklets without sacrificing the overall tracking ratio. 

We validate TAO across diverse deployment conditions in urban simulation environments that combine maps from CARLA~\cite{DosovitskiyEtAl2017}, PTZ cameras implemented in Unreal Engine, and targets modeled as unmanned aerial vehicles (UAVs) governed by AirSim~\cite{ShahEtAl2018} flight dynamics. Across the eight urban maps, TAO achieves the highest PTSD among the evaluated approaches, improving PTSD by $10.5\%$ on average over the six competing baselines. Moreover, the derived potential exhibits broadly similar method-wise trends to PTSD while avoiding simulator-based reward evaluation and reducing the estimated training cost by at least $338.88\times$.


\section{Background and Related Work}
\label{sec:background_relatedwork}

Existing camera-placement research spans geometry-based view selection, graph-based combinatorial optimization, and RL-based sequential decision making. We review these directions below and discuss their relevance to persistent cross-camera tracking.

\paragraph{Camera Placement and View Selection.}
Selecting informative viewpoints is a long-standing problem in visual sensing. Classical methods discretize the environment, evaluate the visibility of each candidate pose, and optimize geometric objectives such as scene coverage, viewing angle, and redundancy. H\"orster and Lienhart~\cite{HorsterLienhart2006} formulated multi-sensor placement as a discrete optimization problem and compared optimal and heuristic solvers, while Hanoun et al.~\cite{HanounEtAl2017} developed a design framework tailored to active PTZ camera networks. Su et al.~\cite{SuEtAl2021} modeled optimal camera placement as a unicost set-covering problem and proposed a weighting-based variable-neighborhood search that combines instance reduction with efficient local swaps. 

Learning-based methods reduce the cost of repeatedly evaluating visibility. Sun et al.~\cite{SunEtAl2021} learned a differentiable surrogate for multi-view coverage and converted discrete view selection into continuous camera-pose optimization. DISeR~\cite{KlinghofferEtAl2023} further demonstrated that RL can jointly design task-dependent imaging systems. More recently, Chen et al.~\cite{ChenEtAl2026} derived the CONVERGE criterion from an approximation of Fisher information gain and showed that informative views preferentially observe geometry that is under-covered by existing cameras. These methods provide increasingly efficient ways to select views for a static scene or reconstruction objective, but they do not explicitly characterize the temporal continuity required to track unpredictable moving targets.

\paragraph{Graph-Based Combinatorial Optimization.}
A convenient representation for deployment is obtained by discretizing candidate camera poses and visible space into two vertex sets. An edge connects a candidate to every voxel visible from that pose, producing a bipartite graph on which selecting a fixed number of cameras becomes a maximum coverage problem (MCP). This representation separates expensive visibility computation from subsequent combinatorial search and naturally supports large candidate sets. Early learning-based graph solvers provided alternatives to handcrafted heuristics. S2V-DQN~\cite{KhalilEtAl2017} combined structure2vec embeddings with an RL agent trained by fitted Q-learning to learn a greedy policy that incrementally constructs graph solutions. The graph convolutional network (GCN)-based method GCN-TreeSearch~\cite{LiEtAl2018} subsequently used supervised predictions to guide a diversified tree search, improving solution quality and generalization to larger graphs; however, it relied on labeled solutions and repeated search.

GCOMB~\cite{ManchandaEtAl2020} then shifted the focus toward budget-constrained optimization on very large graphs by combining probabilistic-greedy node-quality prediction, aggressive candidate pruning, and importance-sampled Q-learning. DIFUSCO~\cite{SunYang2023} later represented a solution as a binary node- or edge-selection vector and generated solutions through graph-based denoising diffusion, allowing parallel refinement and multimodal solution generation. However, its supervised denoising objective assumes that each training instance is paired with an optimal or high-quality solution. Producing such labels requires repeatedly solving the original deployment problem and is therefore impractical when maps, occlusions, target trajectories, and camera budgets vary. Moreover, MCP-based methods primarily optimize spatial coverage; without an explicitly tracking-aware objective, high coverage may still yield fragmented observations of dynamic targets.

\paragraph{Reinforcement Learning.}
The absence of ground-truth deployments makes RL particularly suitable for our setting. Following the standard formulation of Sutton and Barto~\cite{SuttonBarto2018}, sequential decision making in RL is modeled as a Markov decision process (MDP), in which the next state and reward depend only on the current state and action rather than the full interaction history. At step $t$, an agent receives an observation $o_t$ that summarizes the current partial deployment, chooses an action $a_t$ such as adding a feasible camera candidate, receives a reward $r_t$ measuring the resulting improvement, and transitions to the next state. A policy is optimized to maximize the expected cumulative reward, allowing task objectives derived from simulation or analytical surrogates to replace unavailable solution labels.

Within this framework, the deep Q-network (DQN) of Mnih et al.~\cite{MnihEtAl2015} provides a practical value-based method for learning the sequential deployment policy. It approximates the action-value function $Q(o_t,a_t)$ with a neural network and learns from temporal-difference targets of the form $r_t+\gamma\max_{a'}Q(o_{t+1},a')$. Experience replay reduces correlations between successive samples, while a slowly updated target network stabilizes training. For camera deployment, infeasible or previously selected candidates can be masked out, and the policy can construct a solution one camera at a time. This formulation directly connects geometric observations, a tracking-aware reward, and discrete placement actions without requiring an optimal camera configuration as supervision. 

Our method follows this direction but differs from coverage-only graph solvers by optimizing a potential that jointly accounts for target reacquisition and continuous cross-camera tracking. Across eight maps, TAO improves PTSD by $10.5\%$ on average over the six competing baselines, while the low-cost potential preserves broadly similar method-wise trends to PTSD and avoids simulator-based training rollouts.

\section{Approach}
\label{sec:approach}

\begin{figure}
    \centering
    \setbox0=\hbox{\includegraphics[width=\linewidth]{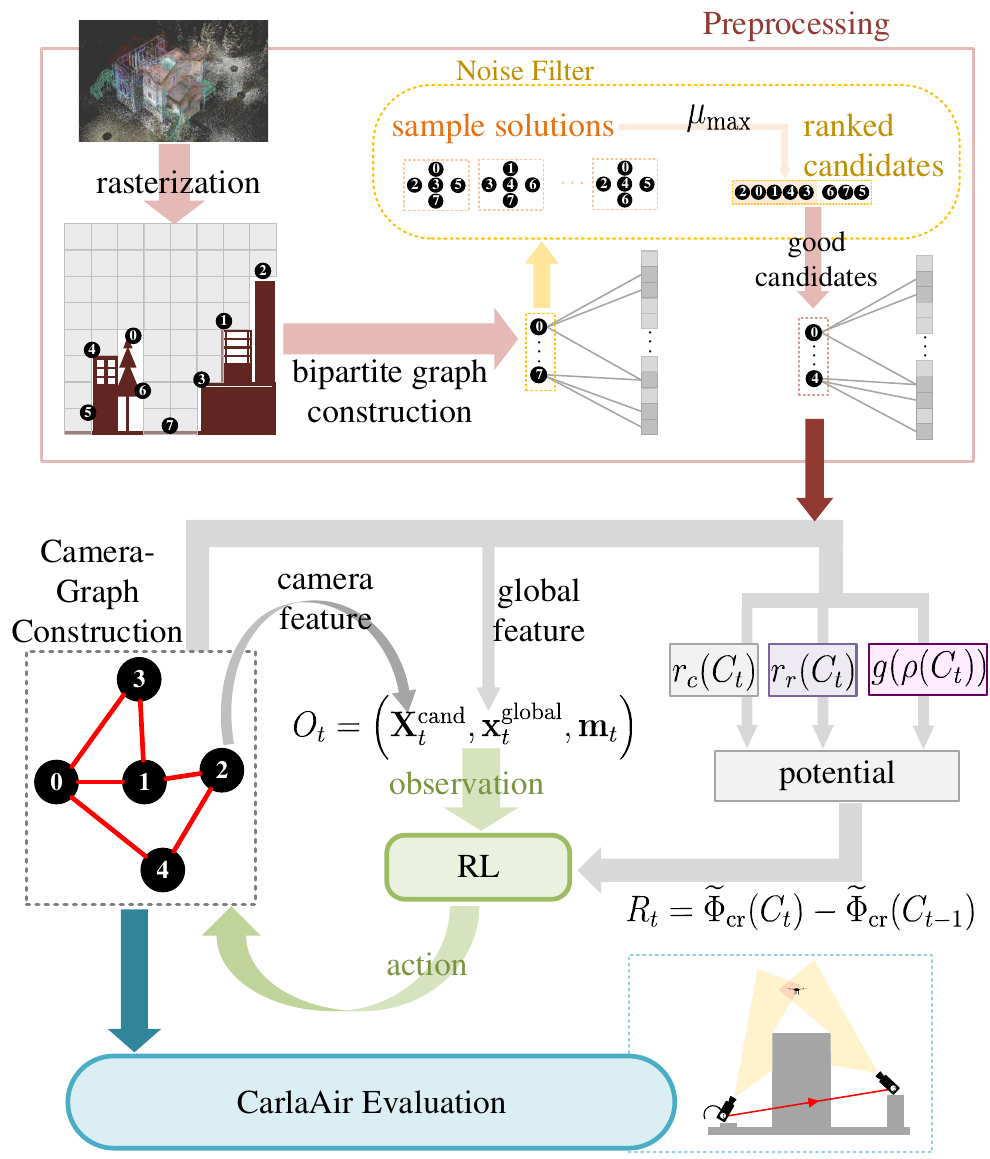}}
    \includegraphics[width=1\linewidth,height=\ht0]{fig/process_pdf.pdf}
    \caption{Overview of the proposed TAO framework.}
    \label{fig:process}
\end{figure}

The proposed TAO consists of three stages: candidate preprocessing, RL-based camera selection, and tracking-aware camera-graph construction. Before presenting these stages, we define the tracking objective and its geometric surrogate. An overview is provided in Figure~\ref{fig:process}.

\subsection{Tracking-Aware Formulation}
\label{sec:approach_formulation}

\paragraph{Problem setup and PTSD.}
We discretize the feasible space\footnote{The feasible space is the collision-free three-dimensional region of a map through which targets can move freely. It can be constructed from geospatial data such as Google Maps, three-dimensional scans, or point clouds reconstructed using simultaneous localization and mapping (SLAM).} into a voxel set
$V=\{v_1,v_2,\ldots\}$ and denote a camera configuration by
$C=\{c_1,c_2,\ldots,c_{|C|}\}$. The set of voxels visible to camera
$c$ is denoted by $V^{[c]}\subseteq V$. Let
$S=\{s_1,s_2,\ldots,s_{|S|}\}$ be a collection of target trajectories,
where the trajectory of target $i$ is the ordered voxel sequence
$s_i=(v_{i,1},v_{i,2},\ldots,v_{i,T_i})$, 
and $v_{i,t}$ denotes the voxel traversed by target $i$ at time step $t$.
For a fixed camera configuration, $s_i$ is partitioned into $M_i$ maximal,
mutually disjoint tracking subtrajectories
$s_{i,1},\ldots,s_{i,M_i}$. We measure trackability using the average PTSD,
defined as
\begin{equation}
    \mathcal{O}(S)
    =\frac{1}{|S|}\sum_{s_i\in S}\frac{1}{|s_i|}
    \sum_{m=1}^{M_i}|s_{i,m}|^2.
    \label{eq:ptsd}
\end{equation}
The quadratic term rewards long tracking segments more strongly than
fragmented segments with the same total temporal span.

\paragraph{Coverage.}
The spatial coverage ratio of configuration $C$ is defined as
\begin{equation}
    r_c(C)=\frac{\left|\bigcup_{c\in C}V^{[c]}\right|}{|V|}.
    \label{eq:coverage_ratio}
\end{equation}
Broad coverage increases the probability that a previously unobserved target
is acquired or reacquired by the camera team.

\paragraph{Robustness.}
Let $V_e^{[c]}$ denote the boundary voxels of camera $c$'s visible region.\footnote{Boundary voxels are defined with respect to $V^{[c]}$: a voxel $v\in V^{[c]}$ belongs to $V_e^{[c]}$ if at least one of its neighboring voxels does not belong to $V^{[c]}$.}
We denote the other cameras in the configuration by
$C\setminus\{c\}$\footnote{For two sets $A$ and $B$, the set difference
$A\setminus B$ contains the elements that belong to $A$ but not to $B$.}, and then quantify deployment robustness by the average boundary overlap
\begin{equation}
    r_r(C)=\frac{1}{|C|}\sum_{c\in C}
    \frac{\left|V_e^{[c]}\cap
    \left(\bigcup_{\ell\in C\setminus\{c\}}V^{[\ell]}\right)\right|}
    {|V_e^{[c]}|}.
    \label{eq:redundancy_ratio}
\end{equation}
This overlap provides neighboring cameras with shared boundary regions in
which target handovers can occur. The resulting one-step continuation
probability can be expressed as
\begin{equation}
    D(C)=\frac{1}{1+\delta_d}
    +\frac{\delta_d\,\tau(|S|,|C|)}{1+\delta_d}r_r(C),
    \label{eq:continuation_probability}
\end{equation}
where $d$ is the distance a target can travel in one time step, $\delta_d$
is the corresponding first-order relative dilation factor of a camera FoV,
and $\tau(|S|,|C|)$ is the fraction of camera capacity available for a
handover under the current workload. We derive Eq.~\eqref{eq:continuation_probability}
in Supplementary Material Sec.~A. Thus, robustness increases the probability of
continuous cross-camera tracking after a target leaves its current FoV.

\paragraph{Potential.}
Directly evaluating PTSD requires online deployment or costly simulation.
We derive the long-horizon limit of expected PTSD under the observation
dynamics induced by $C$ in Supplementary Material Sec.~A and call the resulting
geometric surrogate the \emph{potential}:
\begin{equation}
\begin{split}
    \Phi(C)
    &\triangleq
    \lim_{T\rightarrow\infty}\mathbb{E}[\mathcal{O}(S)\mid C]\\
    &=\frac{r_c(C)}{1-D(C)+r_c(C)}
    \cdot\frac{1+D(C)}{1-D(C)}.
    \label{eq:configuration_potential}
\end{split}
\end{equation}
The potential combines complementary properties of a camera configuration:
$r_c(C)$ supports target acquisition and reacquisition, whereas $r_r(C)$
increases $D(C)$ and promotes continuous cross-camera handovers. It can
therefore be computed inexpensively from deployment geometry while retaining
the tracking-aware structure of PTSD.

\subsection{Preprocessing}
\label{sec:approach_preprocessing}

Given the discretized voxel set of the feasible region $V=\{v_i\}$, we also sample a set of camera-placement candidates $B=\{b_j\}$ from building surfaces, the ground, and other feasible supporting structures. This restriction reflects practical requirements of PTZ cameras, including mechanical support and access to nearby power and communication infrastructure. The voxels and placement candidates are represented as two disjoint vertex sets of a bipartite graph
\begin{equation}
    G_b=(B\cup V,E_b),\qquad
    (b_j,v_i)\in E_b
    \Longleftrightarrow v_i\in V^{[b_j]},
    \label{eq:bipartite_visibility_graph}
\end{equation}
where $V^{[b_j]}$ is the voxel set visible from $b_j$. Thus, every edge indicates that a camera deployed at the corresponding candidate can observe the connected voxel. The purpose of this stage is to remove a large number of low-quality candidates while preserving the candidates most useful to the camera team, thereby substantially reducing the training and inference costs of TAO.

Our pruning procedure follows the probabilistic-greedy node-quality prediction strategy of GCOMB~\cite{ManchandaEtAl2020}, which was designed for large-scale budget-constrained combinatorial optimization, including the MCP on bipartite graphs. We first rank all candidates in descending order of degree. Because $\deg(b_j)=|V^{[b_j]}|$, top-ranked candidates generally provide broad FoVs. We introduce the preprocessing hyperparameter $\mu_{\max}$ to bound the candidate-pool size, and let $B_{\mu_{\max}}\subseteq B$ contain the first $\mu_{\max}$ candidates in this ranking. Degree alone, however, cannot distinguish redundant coverage from coverage of scarce voxels. We therefore sample multiple partial solutions from $B_{\mu_{\max}}$ using probabilistic greedy selection. For a partial set $A\subseteq B_{\mu_{\max}}$, the marginal gain of an unselected candidate is
\begin{equation}
    \Delta(b_j\mid A)
    =\left|V^{[b_j]}\setminus\bigcup_{b\in A}V^{[b]}\right|,
\end{equation}
and the corresponding sampling probability is
\begin{equation}
    \Pr(b_j\mid A)
    =\frac{\Delta(b_j\mid A)}
    {\sum_{b\in B_{\mu_{\max}}\setminus A}\Delta(b\mid A)}.
    \label{eq:probabilistic_greedy_sampling}
\end{equation}
Sampling stops when the normalized marginal gain becomes negligible. Across all sampled solutions, we record the degree rank of every selected candidate and define the adaptive cutoff $\widehat{\mu}$ as the largest rank observed, where $\widehat{\mu}\leq\mu_{\max}$. All candidates in $B_{\mu_{\max}}$ whose ranks do not exceed $\widehat{\mu}$ then form the good-candidate set $B_g$.

In short, this procedure preserves candidates that add otherwise uncovered or rarely covered voxels even when their individual degrees are not maximal. The upper part of Figure~\ref{fig:process} illustrates this preprocessing stage. The pruning procedure determines $|B_g|$, and the RL policy subsequently selects a final deployment of $|C|$ cameras from $B_g$.

\subsection{RL-Based Selection}
\label{sec:approach_rl}

Given $B_{g}$, the second stage selects a subset $C\subseteq B_g$ as the final camera configuration. The policy seeks to maximize the potential in Eq.~\eqref{eq:configuration_potential}, thereby favoring placements that jointly support broad target acquisition and robust continuous handovers.

However, directly optimizing the potential can introduce robustness too early: increasing $r_r$ by placing nearby cameras may sharply reduce $r_c$, leading to a poor team configuration. We address this failure mode with \emph{coverage retention}. For each map and camera budget, we construct the fixed reference $C^{\mathrm{cov}}$ by starting from the empty set and greedily adding the candidate in $B_g$ with the largest marginal coverage gain until $|C|$ cameras have been selected. Thus, $C^{\mathrm{cov}}$ is a coverage-only reference configuration computed from the current candidate set. We then define
\begin{equation}
    \rho(C_t)=\operatorname{clip}
    \left(\frac{r_c(C_t)}{r_c(C^{\mathrm{cov}})},0,1\right).
    \label{eq:coverage_retention}
\end{equation}
The contribution of robustness is then controlled by a smooth gate
\begin{equation}
g(\rho)=
\begin{cases}
0, & \rho\leq\rho_0,\\
3u^2-2u^3, & \rho_0<\rho<1,\\
1, & \rho\geq1,
\end{cases}
\label{eq:coverage_retention_gate}
\end{equation}
where $u=\frac{\rho-\rho_0}{1-\rho_0}$ and $\rho_0$ is the gating threshold.
Specifically, TAO replaces $r_r(C_t)$ in $\Phi(C_t)$ with the effective robustness $\widehat{r}_r(C_t)=g(\rho(C_t))r_r(C_t)$ and correspondingly replaces $\Phi(C_t)$ with $\widetilde{\Phi}(C_t)$, which denotes the resulting coverage-retained potential normalized by its attainable upper bound. Consequently, robustness cannot dominate policy learning until the current configuration retains nearly all of the coverage provided by the coverage-only reference, while the smooth transition avoids a discontinuity in the learning signal. Robustness and coverage retention therefore work together to extend continuous tracking duration without sacrificing overall trackability.

With the coverage-retained potential defined, we next specify the RL learning signal and decision process. The stepwise reward is
\begin{equation}
    R_t=\widetilde{\Phi}(C_t)
    -\widetilde{\Phi}(C_{t-1}),
    \label{eq:tao_reward}
\end{equation}
which provides a dense learning signal and telescopes over an episode to the change between the final and initial normalized potentials. The observation is defined by
\begin{equation}
    O_t=\left(\mathbf{X}^{\mathrm{cand}}_t,
    \mathbf{x}^{\mathrm{global}}_t,\mathbf{m}_t\right),
    \label{eq:tao_observation}
\end{equation}
where $\mathbf{X}^{\mathrm{cand}}_t$ contains candidate-level features such as the selected flag, marginal coverage and gated-robustness gains, post-selection coverage, coverage-retention margin, and summary statistics of continuous coverage, blind gaps, and redundant coverage. The global vector $\mathbf{x}^{\mathrm{global}}_t$ describes the numbers of used and remaining cameras, the current $r_c$ and $r_r$, their interaction, the retention ratio, and aggregate boundary-level statistics. The action mask $\mathbf{m}_t$ excludes candidates already selected. Given the observation $O_t$, the RL policy chooses an action $a_t\in B_g\setminus C_{t-1}$ that selects one previously unselected camera candidate and updates the partial configuration as $C_t=C_{t-1}\cup\{a_t\}$; the resulting transition is evaluated by the reward $R_t$. 
Starting from $C_0=\varnothing$, this process continues until $|C_t|=|C|$, at which point the episode terminates. Because every action adds exactly one camera, each episode contains exactly $|C|$ steps and produces a final configuration of exactly $|C|$ cameras.

\subsection{Tracking-Aware Camera-Graph Construction}
\label{sec:approach_camera_graph}

After a camera configuration $C$ has been selected, we automatically construct an undirected camera graph $\mathcal{G}_{C}=(C,\mathcal{E}_{C})$. Each camera is a vertex, and two cameras are connected when their visible voxel sets overlap:
\begin{equation}
    (c_i,c_j)\in\mathcal{E}_{C}
    \Longleftrightarrow
    V^{[c_i]}\cap V^{[c_j]}\neq\varnothing.
    \label{eq:camera_graph}
\end{equation}
Every edge is a bidirectional communication link. When a target approaches the boundary of one camera's FoV and moves through the overlapping voxels toward a neighboring camera, a handover message is sent from the former camera to the latter along this edge. This distributed tracking-aware mechanism requires only local communication between cameras with overlapping FoVs and therefore avoids network-wide broadcasting. Moreover, overlapping cameras are typically spatially close, which keeps communication latency and signal attenuation manageable.

Through this graph construction, the camera team is explicitly endowed with tracking awareness: coverage supports target acquisition and reacquisition, while overlap-aware communication supports continuous cross-camera handovers. The experiments in Sec.~\ref{sec:results} demonstrate that the complete TAO pipeline substantially extends continuous tracking duration without sacrificing the overall tracking ratio. Further analyses of handover performance are provided in Supplementary Material Sec.~B.

\section{Results}
\label{sec:results}
\setcounter{topnumber}{1}

Using UAVs as dynamic targets, we conduct extensive simulations on CARLA urban
maps and demonstrate the advantages of the proposed framework through
comparisons with representative methods and targeted ablation studies.

\subsection{Experimental Settings}
\label{sec:experimental_settings}

We construct the simulation environment with CARLA-Air~\cite{ZengEtAl2026}, which integrates urban environments from CARLA~\cite{DosovitskiyEtAl2017} and UAV dynamics from AirSim~\cite{ShahEtAl2018} within a unified Unreal Engine runtime. Eight CARLA maps are used: Town01--Town07 and Town10HD. TAO is trained only on Town01 and is directly evaluated, \emph{without further fine-tuning}, on all eight maps. This protocol tests not only in-domain performance but also cross-map generalization to previously unseen urban layouts. For each map, we consider multi-target tasks containing $1$--$5$ UAVs. At the beginning of each trial, the UAVs depart from the map center in different initial directions and randomly plan their flight trajectories while performing dynamic obstacle avoidance. To account for different map scales, their flight speed is set to $v=\min\{120,\max\{3,\max(W,H)/8\}\}\,\mathrm{m/s}$, where $W$ and $H$ are the usable horizontal spans of the map.

Camera-placement candidates are sampled from feasible supporting surfaces at a spatial interval of $1\,$m, while the surveillance region is discretized into voxels with a side length of $5\,$m. Each PTZ camera has a maximum sensing distance of $100\,$m, a horizontal pan range of $[-180^{\circ},180^{\circ}]$, and a pitch range of $[-75^{\circ},75^{\circ}]$. The deployment budget $|C|$ is varied from 6 to 10. During preprocessing, we set the maximum candidate-pool size to $\mu_{\max}=256$, sample 30 probabilistic-greedy solutions from $B_{\mu_{\max}}$, and stop extending a sampled solution when its normalized marginal gain falls below $\delta=0.01$. As described in Sec.~\ref{sec:approach_preprocessing}, $\mu_{\max}$ is a fixed preprocessing hyperparameter. For each map, the adaptive cutoff $\widehat{\mu}\leq\mu_{\max}$ is the largest degree rank recorded among the selected candidates in the sampled solutions, and candidates with rank at most $\widehat{\mu}$ form $B_g$.
Table~\ref{tab:map_statistics} summarizes the horizontal dimensions and discretization statistics of the eight maps.

\begin{table}[t]
    \centering
    \caption{Statistics of the eight CARLA maps used in the experiments. Map
    size reports the horizontal boundary span; voxel and surface-candidate
    counts are obtained using discretization intervals of $5$ m and $1$ m,
    respectively.}
    \label{tab:map_statistics}
    \resizebox{\columnwidth}{!}{%
    \begin{tabular}{lccc}
        \toprule
        Map & \shortstack{Map size\\($W\times H$, m)} &
        \shortstack{Number of\\voxels} &
        \shortstack{Number of surface\\candidates} \\
        \midrule
        Town01   & $330\times330$ & 24,109  & 46,803 \\
        Town02   & $200\times200$ & 7,757   & 13,168 \\
        Town03   & $400\times400$ & 32,139  & 67,510 \\
        Town04   & $932\times834$ & 179,614 & 21,557 \\
        Town05   & $350\times350$ & 27,415  & 44,487 \\
        Town06   & $900\times260$ & 50,318  & 20,487 \\
        Town07   & $250\times300$ & 16,435  & 11,231 \\
        Town10HD & $200\times200$ & 6,727   & 3,170  \\
        \bottomrule
    \end{tabular}%
    }
\end{table}

Because the representative RL-based baselines S2V-DQN and GCOMB both use DQN architectures, we instantiate TAO with the same architecture for a fair comparison.
We train the resulting TAO policy for 10,000 environment steps with a learning rate of $10^{-5}$, a discount factor of $0.99$, a batch size of 32, and a replay-buffer capacity of 40,000. The action-value model contains two branches: a contextual feature-weight branch and a state-value branch. Each branch is a multilayer perceptron with one 128-unit hidden layer and a rectified linear unit (ReLU) activation. The coverage and robustness terms have equal weights of 0.5. For coverage retention, $C^{\mathrm{cov}}$ is recomputed once for every map and camera budget using the coverage-only greedy procedure in Sec.~\ref{sec:approach_rl}; it is a derived reference configuration and introduces no tunable value. The smooth gate in Eq.~\eqref{eq:coverage_retention_gate} uses the hyperparameter $\rho_0=0.95$, so robustness begins to affect the objective only after most of the coverage-only reference has been retained. Qualitative examples of TAO deployments on all eight CARLA maps, together with UAV trajectories and the corresponding coverage and robustness voxels, are provided in Supplementary Material Sec.~C.

\begin{table}[t]
  \centering
  \caption{Overall tracking performance across all evaluation settings.}
  \label{tab:main_table}
  \small
  \begin{tabular}{@{}llll@{}}
    \toprule
    Method & PTSD $\uparrow$ & MaxD $\uparrow$ & TR $\uparrow$ \\
    \midrule
    Random & 200.163 & 441.965 & 0.4507 \\
    Beam Search & 293.707 & 528.400 & 0.5658 \\
    Greedy & 311.443 & 547.350 & 0.5789 \\
    GBFS & 312.095 & \textbf{549.000} & 0.5791 \\
    \midrule
    S2V-DQN & 299.139 & 526.600 & 0.5802 \\
    GCOMB & 302.966 & 534.750 & 0.5743 \\
    TAO (Ours) & \textbf{316.542} & 548.050 & \textbf{0.5814} \\
    \bottomrule
  \end{tabular}
\end{table}

\subsection{Evaluation Metrics}
\label{sec:evaluation_metrics}

We report three metrics, for which higher values indicate better performance.
Cross-camera handovers involve target recognition and observation assignment,
which are implementation details orthogonal to camera placement. We therefore
do not require a strictly zero interruption between observations from
consecutive cameras: two intervals separated by no more than 30 steps are
treated as the same tracking chain, whereas a longer interruption starts a new
chain. Each $s_{i,m}$ below denotes one such maximal team-level chain and spans
from its first to its last observed step. PTSD is computed using
Eq.~\eqref{eq:ptsd}, jointly measuring the fraction of the trajectory spanned
by tracking chains and their temporal persistence. Let
\begin{equation}
    s_i^{\text{max}}=\max_{1\leq m\leq M_i}|s_{i,m}|,
\end{equation}
where $s_i^{\text{max}}=0$ if target trajectory $s_i$ is never tracked. The
maximum continuous tracking duration (MaxD) is
\begin{equation}
    \mathrm{MaxD}=\max_{s_i\in S}s_i^{\text{max}},
    \label{eq:duration_metrics}
\end{equation}
    and is measured in simulation steps. Finally, the tracking ratio (TR) is
\begin{equation}
    \mathrm{TR}=\frac{1}{|S|}\sum_{s_i\in S}
    \frac{\sum_{m=1}^{M_i}|s_{i,m}|}{|s_i|},
    \label{eq:track_ratio}
\end{equation}
which is the mean fraction of each target trajectory covered by tracking chains.

\subsection{End-to-End Performance}
\label{sec:main_comparison}

We introduce six baselines for comparison. Among the four heuristic approaches, Random uniformly samples a deployment configuration from the feasible candidates. Beam Search~\cite{FurcyKoenig2005} retains a fixed-width set of the best partial configurations at each search depth, trading search breadth for bounded memory.
\begin{table}[t]
  \centering
  \caption{Ablation study across all evaluation settings.}
  \label{tab:ablation_table}
  \small
  \begin{tabular}{@{}llll@{}}
    \toprule
    Method & PTSD $\uparrow$ & MaxD $\uparrow$ & TR $\uparrow$ \\
    \midrule
    TAO & \textbf{316.542} & \textbf{548.050} & \textbf{0.5814} \\
    \midrule
    w/o robustness & 312.163 & 544.050 & 0.5794 \\
    \midrule
    \shortstack[l]{w/o coverage\\retention} & 147.700 & 333.550 & 0.3417 \\
    \bottomrule
  \end{tabular}
\end{table}

Greedy~\cite{NemhauserEtAl1978} repeatedly chooses the candidate with the largest immediate marginal gain, a standard strategy for monotone submodular coverage objectives. GBFS~\cite{Pearl1984} maintains a global priority queue of partial configurations ordered by their objective values.

For the two learning-based approaches, S2V-DQN~\cite{KhalilEtAl2017} obtains candidate representations through iterative structure2vec message passing on the candidate--voxel bipartite graph, while GCOMB~\cite{ManchandaEtAl2020} applies a one-hop graph convolution after candidate pruning. TAO directly scores the explicit candidate-level and global observations in Eq.~\eqref{eq:tao_observation}, including marginal coverage, robustness, coverage retention, blind-gap, and handover-continuity features. To isolate the effect of the observation representation, all three methods use exactly the same action space and TAO reward in Eq.~\eqref{eq:tao_reward}. They differ in how they represent and score the current partial deployment. The comparison therefore tests whether TAO's explicit tracking-aware observation design provides a more effective policy input than the generic graph encoders under an identical optimization objective. For evaluation, each cell in Table~\ref{tab:main_table} averages $8$ maps $\times 5$ camera budgets $\times 5$ UAV counts $=200$ runs.

As shown in Table~\ref{tab:main_table}, TAO achieves the highest PTSD
 and TR among all learning-based and heuristic-search methods, while also
 attaining the best MaxD among the learning-based methods. In terms of PTSD, TAO
improves upon the strongest RL-based baseline, GCOMB, by 4.5\% and the
 strongest search baseline, GBFS, by 1.4\%. GBFS obtains a slightly larger MaxD
 than TAO, whereas TAO's higher PTSD and TR indicate a better overall balance
 between tracking-chain duration and actual observation coverage. Because the learned methods
use the same reward and action space, their comparison supports the
effectiveness of TAO's tracking-aware observation design across different
urban layouts and numbers of dynamic targets.

Although PTSD provides a comprehensive measure of tracking performance,
it does not directly expose the quality of camera-to-camera handovers. We
therefore further analyze handover quality using six complementary simulation
metrics that capture successful transitions, post-handover persistence,
repeated handovers, and camera participation. Their definitions, method-wise
comparisons, and component ablations are provided in Supplementary Material Sec.~B.

\begin{table}[t]
  \centering
  \caption{Training-cost comparison over five camera budgets.}
  \label{tab:potential_training_cost}
  \small
  \begin{tabular}{@{}lccc@{}}
    \toprule
    Method & \shortstack{Direct PTSD\\cost (h)} &
    \shortstack{Actual training\\time (min)} & Speedup \\
    \midrule
    S2V-DQN & 1030.49 & 166.90 & $370.45\times$ \\
    GCOMB   & 1086.48 & 160.48 & $406.20\times$ \\
    TAO     & 1088.68 & 192.76 & $338.88\times$ \\
    \bottomrule
  \end{tabular}
\end{table}

\subsection{Ablation Study}
\label{sec:ablation_study}

To isolate the contribution of the two TAO components, we conduct two ablations. \emph{TAO w/o robustness} removes $r_r(C_t)$ from potential and therefore considers coverage alone. \emph{TAO w/o coverage retention} keeps the robustness term but disables $\rho(C_t)$ and sets $g(\cdot)\equiv 1$, allowing robustness to influence the policy from the beginning of an episode.

Table~\ref{tab:ablation_table} demonstrates the effectiveness of both
proposed components. In terms of PTSD, the complete TAO improves upon the
variants without robustness and without coverage retention by 1.4\% and
114.3\%, respectively. Without robustness, the deployment no longer accounts
for overlap that supports cross-camera tracking, reducing both chain duration
and trackability. Without coverage retention, robustness influences the policy
too early and concentrates cameras within a small region, sharply reducing all
three metrics. Its performance is even worse than Random, which
selects cameras uniformly over the feasible candidates.

\subsection{Computational Efficiency of Potential}
\label{sec:potential_efficiency}

We quantify the computational advantage of the potential over direct PTSD-based rewards. Direct PTSD evaluation requires launching the high-fidelity simulator and executing a tracking rollout for every completed camera configuration, whereas the potential is computed from the preprocessed visibility graph without CARLA or AirSim during training.

\begin{figure}[t]
  \centering
  \includegraphics[width=\columnwidth]{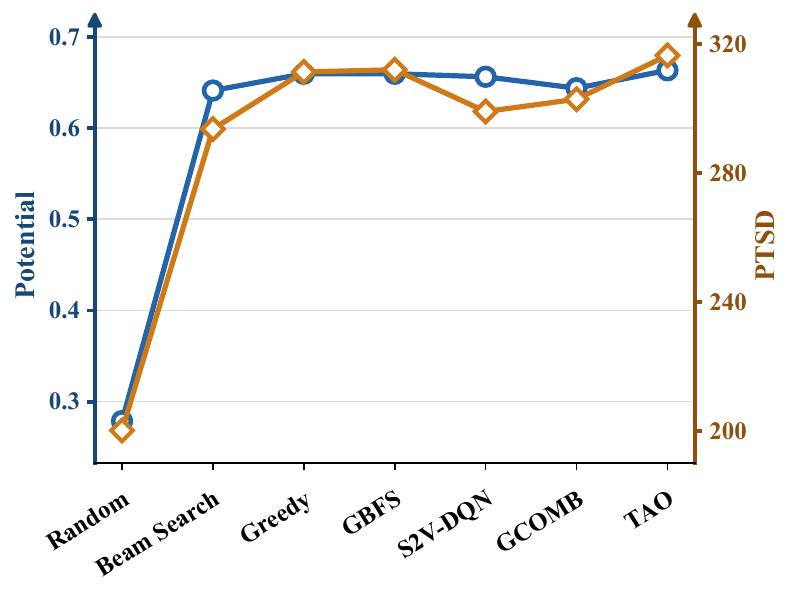}
  \caption{Training-map potential and mean PTSD across all eight evaluation maps.}
  \label{fig:potential_vs_ptsd}
\end{figure}

To estimate this cost, we measure the wall-clock time of the PTSD evaluations and denote the method-specific mean simulation time by $\bar{t}_{\mathrm{sim}}$. On a server equipped with four NVIDIA GeForce RTX 4090 GPUs, $\bar{t}_{\mathrm{sim}}$ is 574.44, 605.65, and 606.88 s for S2V-DQN, GCOMB, and TAO, respectively. Each value is averaged over 200 evaluations covering eight maps, five camera budgets, and five UAV counts. Across budgets $|C|=6,\ldots,10$, the five training runs contain 6,458 complete deployments. We therefore estimate the direct-PTSD cost as $T_{\mathrm{direct}}=6{,}458\,\bar{t}_{\mathrm{sim}}/3{,}600$ GPU-hours, while the reported actual training time is the sum of the five potential-based runs on Town01.

Table~\ref{tab:potential_training_cost} shows that direct PTSD estimation would require 1,030.49--1,088.68 simulator GPU-hours per method, compared with 160.48--192.76 minutes (i.e., 2.67--3.21 hours) of actual potential-based training. TAO takes 192.76 minutes, slightly longer than S2V-DQN and GCOMB because its reward computation includes tracking-aware robustness and coverage-retention terms. This \emph{one-time} offline overhead does not alter deployment-time DQN action selection. More importantly, the potential avoids simulator rollouts and reduces the estimated training cost by $338.88\times$--$406.20\times$ across the three learning-based methods.

\subsection{PTSD Consistency}
\label{sec:ptsd_consistency}

Figure~\ref{fig:potential_vs_ptsd} compares the potential computed on the Town01 training map with PTSD averaged over all eight evaluation maps. Although the two measures differ in numerical scale and map scope and do not yield identical rankings, they exhibit broadly similar method-wise trends: a higher training-map potential generally corresponds to a higher cross-map PTSD. Together with the speedups in Sec.~\ref{sec:potential_efficiency}, this result supports the potential as a useful, though imperfect, surrogate when repeated high-fidelity PTSD evaluations are prohibitively expensive.

\section{Conclusion}
\label{sec:conclusion}

This paper addressed multi-PTZ-camera placement for the continuous tracking of dynamic 3D targets in complex urban environments. We introduced PTSD, which jointly measures trajectory coverage and tracking duration, and derived the potential as a low-cost geometric surrogate that combines spatial coverage with deployment robustness. TAO optimizes this surrogate using an RL policy whose dense reward incorporates robustness and coverage retention. Experiments across eight urban maps show that TAO achieves the highest PTSD and TR among the evaluated methods, remains competitive in maximum continuous tracking duration (MaxD), and reduces the estimated training cost by $338.88\times$--$406.20\times$ through potential-based learning. The ablations further confirm the complementary benefits of robustness and coverage retention.

\clearpage
\section{LLM Usage Statement}
\label{sec:llm_statement}

We used GPT-5.6 Sol (hereafter, the LLM) to assist with language polishing, primarily in the Background and Related Work section and the Approach section. This assistance included improving word choice and correcting grammatical errors. All mathematical formulations were written manually; during the theoretical derivations, the LLM was used only to check the correctness of the derivation steps. All cited papers were identified and selected manually; the LLM was used only to help correct reference formatting.

The experimental data and visualizations were generated primarily in the CARLA environment, with no LLM involved in their generation. The pipeline diagram was manually created in Microsoft Visio without LLM assistance.

After completing the experiments, we used the LLM to write statistical scripts for computing the mean of each metric in every table cell. Regarding code development, the core algorithmic pipeline was implemented manually, whereas batch-processing shell scripts (\texttt{.sh}) were generated by the LLM.

{
    \small
    \bibliographystyle{ieeenat_fullname}
    \bibliography{main}
}

\clearpage
\setcounter{page}{1}
\renewcommand{\thepage}{S\arabic{page}}
\maketitlesupplementary

\setcounter{section}{0}
\renewcommand{\thesection}{\Alph{section}}
\renewcommand{\thesubsection}{\thesection.\arabic{subsection}}
\renewcommand{\theHsection}{supp.\Alph{section}}
\renewcommand{\theHsubsection}{\theHsection.\arabic{subsection}}
\setcounter{equation}{0}
\renewcommand{\theequation}{S\arabic{equation}}
\renewcommand{\theHequation}{supp.\arabic{equation}}
\setcounter{table}{0}
\renewcommand{\thetable}{\arabic{table}}
\renewcommand{\theHtable}{supp.\arabic{table}}
\setcounter{figure}{0}
\renewcommand{\thefigure}{\arabic{figure}}
\renewcommand{\theHfigure}{supp.\arabic{figure}}

\section{Derivation of the Potential}
\label{sec:potential_derivation}

This section derives the potential in Eq.~\eqref{eq:configuration_potential}
from the expected PTSD under a homogeneous observation model. The notation
and geometric quantities follow Sec.~\ref{sec:approach_formulation}.

\begin{table*}[t]
    \centering
    \caption{Handover performance of the seven methods in the main comparison and
    two TAO ablations over eight maps, five camera budgets, and five UAV counts.}
    \label{tab:handover_details}
    \resizebox{\textwidth}{!}{%
    \begin{tabular}{lcccccc}
        \toprule
        Method & \shortstack{Handover\\rate (\%) $\uparrow$} &
        \shortstack{Post-handover\\duration $\uparrow$} &
        \shortstack{Handovers\\per chain $\uparrow$} &
        \shortstack{Multi-handover chains\\per evaluation $\uparrow$} &
        \shortstack{Distinct cameras\\per chain $\uparrow$} &
        \shortstack{Multi-handover\\chain duration $\uparrow$} \\
        \midrule
        Random & 21.73 & 162.91 & 1.16 & 0.21 & 2.15 & 501.14 \\
        Beam Search & 28.42 & 201.12 & 1.30 & 0.41 & 2.29 & 637.02 \\
        Greedy & 30.26 & 205.09 & 1.34 & 0.47 & 2.33 & 661.90 \\
        GBFS & 30.47 & 206.97 & 1.34 & 0.47 & 2.33 & 665.13 \\
        \midrule
        S2V-DQN & 27.53 & 229.94 & 1.39 & 0.45 & 2.35 & 698.67 \\
        GCOMB & 29.07 & 208.53 & 1.32 & 0.40 & 2.31 & 695.71 \\
        TAO (Ours) & 28.14 & 231.48 & 1.42 &
        0.45 & 2.39 & 713.11 \\
        \midrule
        TAO w/o robustness & 26.78 & 237.47 & 1.44 &
        0.44 & 2.41 & 700.25 \\
        TAO w/o coverage retention & 10.10 & 200.36 & 1.13 &
        0.05 & 2.00 & 344.00 \\
        \bottomrule
    \end{tabular}%
    }
\end{table*}

\subsection{Observation Dynamics}

Let $x_{i,t}\in\{0,1\}$ indicate whether target $i$ belongs to an active tracking chain at time step $t$, and define
\begin{equation}
    p_{i,t}\triangleq P_o(v_{i,t})=\Pr(x_{i,t}=1).
\end{equation}
We further use $P_o(v_{i,t:k})$ to denote the probability that the target is
observed continuously from time $t$ through $k$.

When the target was unobserved at time $t-1$, we approximate its probability
of entering the covered region at time $t$ by
\begin{equation}
    \Pr(x_{i,t}=1,x_{i,t-1}=0)
    \approx (1-p_{i,t-1})r_c.
    \label{eq:observation_entry}
\end{equation}
Conditioned on the target being observed at time $t-1$, its observation can
either persist within the current FoV or be handed over to another camera.
We define the corresponding one-step continuation probability as
\begin{equation}
\begin{split}
    D(C)
    &\triangleq \Pr(x_{i,t}=1\mid x_{i,t-1}=1)\\
    &\approx \frac{1}{1+\delta_d}
    +\frac{\delta_d}{1+\delta_d}\frac{1}{|C|}
    \sum_{c\in C}P_{o,c}^{\mathrm{out}}(v_{i,t}).
\end{split}
\label{eq:observation_continuation}
\end{equation}
Here, $P_{o,c}^{\mathrm{out}}(v_{i,t})$ denotes the probability of a
successful handover after the target leaves the FoV of camera $c$. The first
term represents continued observation within the current FoV, whereas the
second term represents continuation through an inter-camera handover.

The dilation distance $d$ is chosen as the distance a target can travel in
one time step. Let $V_d^{[c]}$ denote the region obtained by dilating the visible region outward $V^{[c]}$ by $d$. The relative dilation factor
$\delta_{c,d}$ is the newly added volume normalized by the original visible
volume:
\begin{equation}
    \delta_{c,d}\triangleq
    \frac{\operatorname{Vol}(V_d^{[c]})-
    \operatorname{Vol}(V^{[c]})}{\operatorname{Vol}(V^{[c]})}.
    \label{eq:relative_dilation_factor}
\end{equation}
This ratio provides a geometric proxy for the likelihood that a target moves
out of the current FoV in one time step. More precisely, the fraction of the
dilated region lying outside the original FoV is
$\delta_{c,d}/(1+\delta_{c,d})$, which is used as the one-step exit
probability in Eq.~\eqref{eq:observation_continuation}.

Let $\operatorname{Sur}(V^{[c]})$ denote the surface area of the boundary of the visible region of
camera $c$. In our voxelized representation, $\operatorname{Sur}(V^{[c]})$ is estimated from the
exposed faces of the boundary voxels $V_e^{[c]}$ defined in
Sec.~\ref{sec:approach_formulation}. Following Weyl~\cite{Weyl1939} and Federer~\cite{Federer1959}, we use a
first-order Minkowski-dilation approximation,
$\operatorname{Vol}(V_d^{[c]})-\operatorname{Vol}(V^{[c]})\approx \operatorname{Sur}(V^{[c]}) d$,
and therefore
$\delta_{c,d}\approx[\operatorname{Sur}(V^{[c]})/\operatorname{Vol}(V^{[c]})]d$. The coefficient
$\operatorname{Sur}(V^{[c]})/\operatorname{Vol}(V^{[c]})$ is the surface-to-volume ratio of the
camera's viewing frustum. Because all cameras share the same nominal viewing-frustum geometry,
we approximate this ratio using the common nominal frustum and therefore treat
it as camera-independent, simplifying $\delta_{c,d}$ to $\delta_d\propto d$.

For camera $c$, we approximate the handover probability as
\begin{equation}
    P_{o,c}^{\mathrm{out}}(v_{i,t})
    \approx
    \frac{\left|V_e^{[c]}\cap
    \left(\bigcup_{\ell\in C\setminus\{c\}}V^{[\ell]}\right)\right|}
    {|V_e^{[c]}|}\,\tau(|S|,|C|),
    \label{eq:handover_probability}
\end{equation}
where $\tau(|S|,|C|)$ is the fraction of camera capacity available for a
handover under a workload of $|S|$ trajectories. Averaging this probability
over cameras and applying the definition of $r_r(C)$ in
Eq.~\eqref{eq:redundancy_ratio} gives
\begin{equation}
    \frac{1}{|C|}\sum_{c\in C}P_{o,c}^{\mathrm{out}}(v_{i,t})
    \approx \tau(|S|,|C|)r_r(C).
    \label{eq:average_handover_probability}
\end{equation}
Substituting Eq.~\eqref{eq:average_handover_probability} into
Eq.~\eqref{eq:observation_continuation} yields $D(C)$ in
Eq.~\eqref{eq:continuation_probability}. For brevity, we write $D=D(C)$
below.

Combining the entry and continuation events gives
\begin{equation}
\begin{split}
    p_{i,t}
    &=r_c(1-p_{i,t-1})+Dp_{i,t-1}\\
    &=(D-r_c)p_{i,t-1}+r_c.
\end{split}
\label{eq:observation_recurrence}
\end{equation}
With $p_{i,1}=r_c$, the solution is
\begin{equation}
\begin{split}
    p_{i,t}&=p_\infty+(r_c-p_\infty)(D-r_c)^{t-1},\\
    p_\infty&=\frac{r_c}{1-D+r_c},
\end{split}
    \label{eq:observation_closed_form}
\end{equation}
provided that $|D-r_c|<1$.

\subsection{Expected PTSD and Its Long-Horizon Limit}

For the resulting binary tracking-chain sequence, the squared lengths of all
contiguous active runs satisfy
\begin{equation}
    \sum_{m=1}^{M_i}|s_{i,m}|^2
    =\sum_{t=1}^{T_i}x_{i,t}
    +2\sum_{1\leq t<k\leq T_i}\prod_{h=t}^{k}x_{i,h}.
    \label{eq:run_length_identity}
\end{equation}
Assume that all trajectories have the same length $T$ and follow the
homogeneous transition model above. Since
$P_o(v_{i,t:k})\approx p_{i,t}D^{k-t}$, taking the expectation of
Eq.~\eqref{eq:run_length_identity} gives
\begin{equation}
\begin{split}
    \mathbb{E}[\mathcal{O}(S)\mid C]
    &=\frac{1}{T}\left[
    \sum_{t=1}^{T}p_{i,t}
    +2\sum_{1\leq t<k\leq T}p_{i,t}D^{k-t}
    \right]\\
    &=\frac{1}{T}\sum_{t=1}^{T}p_{i,t}
    \left(1+2\sum_{h=1}^{T-t}D^h\right).
\end{split}
\label{eq:expected_ptsd}
\end{equation}
Substituting Eq.~\eqref{eq:observation_closed_form} into
Eq.~\eqref{eq:expected_ptsd} produces the finite-horizon expression
\begin{equation}
\begin{split}
    \mathbb{E}[\mathcal{O}(S)\mid C]
    &=\frac{1}{T}\sum_{t=1}^{T}
    \Big[p_\infty+(r_c-p_\infty)(D-r_c)^{t-1}\Big]
    \\
    &\quad\times
    \left(1+2\sum_{h=1}^{T-t}D^h\right).
\end{split}
\label{eq:expected_ptsd_closed_form}
\end{equation}
For sufficiently long trajectories and $|D|<1$, the transient term
vanishes and the geometric sum converges. Hence,
\begin{equation*}
    \lim_{T\rightarrow\infty}\mathbb{E}[\mathcal{O}(S)\mid C]
    =\frac{r_c}{1-D+r_c}\cdot\frac{1+D}{1-D},
\end{equation*}
which is the definition of the potential $\Phi(C)$ in Eq.~\eqref{eq:configuration_potential}.
The derivation also makes its two components explicit: coverage $r_c$
controls acquisition and reacquisition, while robustness $r_r$ increases
$D$ and promotes continuous cross-camera handovers.


\section{Handover Details}
\label{sec:handover_details}

We further analyze inter-camera handovers for every UAV.
Table~\ref{tab:handover_details} reports six handover metrics: handover rate,
post-handover duration, handovers per chain, multi-handover chains per
evaluation, distinct cameras per chain, and multi-handover chain duration. Each observation interval that ends before the simulation terminates is
treated as a handover opportunity. The handover rate is the fraction of these
opportunities followed by a successful cross-camera continuation under the
criterion in Sec.~\ref{sec:evaluation_metrics}. All reported rates pool
opportunities across the 200 evaluations of each method or ablation variant. For each successful
handover, the post-handover duration begins when the receiving camera starts
observing the target and ends when the corresponding team-level tracking chain
terminates under the same criterion. A handover \emph{chain} is a team-level tracking episode containing at least one successful handover. Handovers per chain is the mean number of successful handovers within such chains. A
multi-handover chain contains at least two successful handovers. The number of multi-handover chains per evaluation is the mean number of such chains in one simulation.
Distinct cameras per chain is the mean number of unique cameras
participating in a handover chain, while multi-handover chain duration is the
mean span of multi-handover chains. All durations are reported in simulation
steps.

Among the seven methods in the main comparison, TAO ranks first on four of the six measures in Table~\ref{tab:handover_details} and ties for second on one other measure. The ablations further clarify the effects of
its two components. Removing robustness reduces the handover rate from
28.14\% to 26.78\%, multi-handover chains per evaluation from 0.45 to 0.44,
and multi-handover chain duration from 713.11 to 700.25 steps. Although this
variant has slightly larger conditional means for post-handover duration,
handovers per chain, and distinct cameras per chain, these quantities are
computed only from chains that already contain a successful handover and do
not indicate stronger overall handover performance. Removing coverage
retention degrades all six measures, including a handover rate of 10.10\% and
a multi-handover chain duration of only 344.00 steps. Overall, the complete
design provides the strongest balance between handover frequency, repeated
transitions, and long-term persistence.

\begin{figure*}[t]
    \centering
    \setbox0=\hbox{\includegraphics[width=0.75\linewidth]{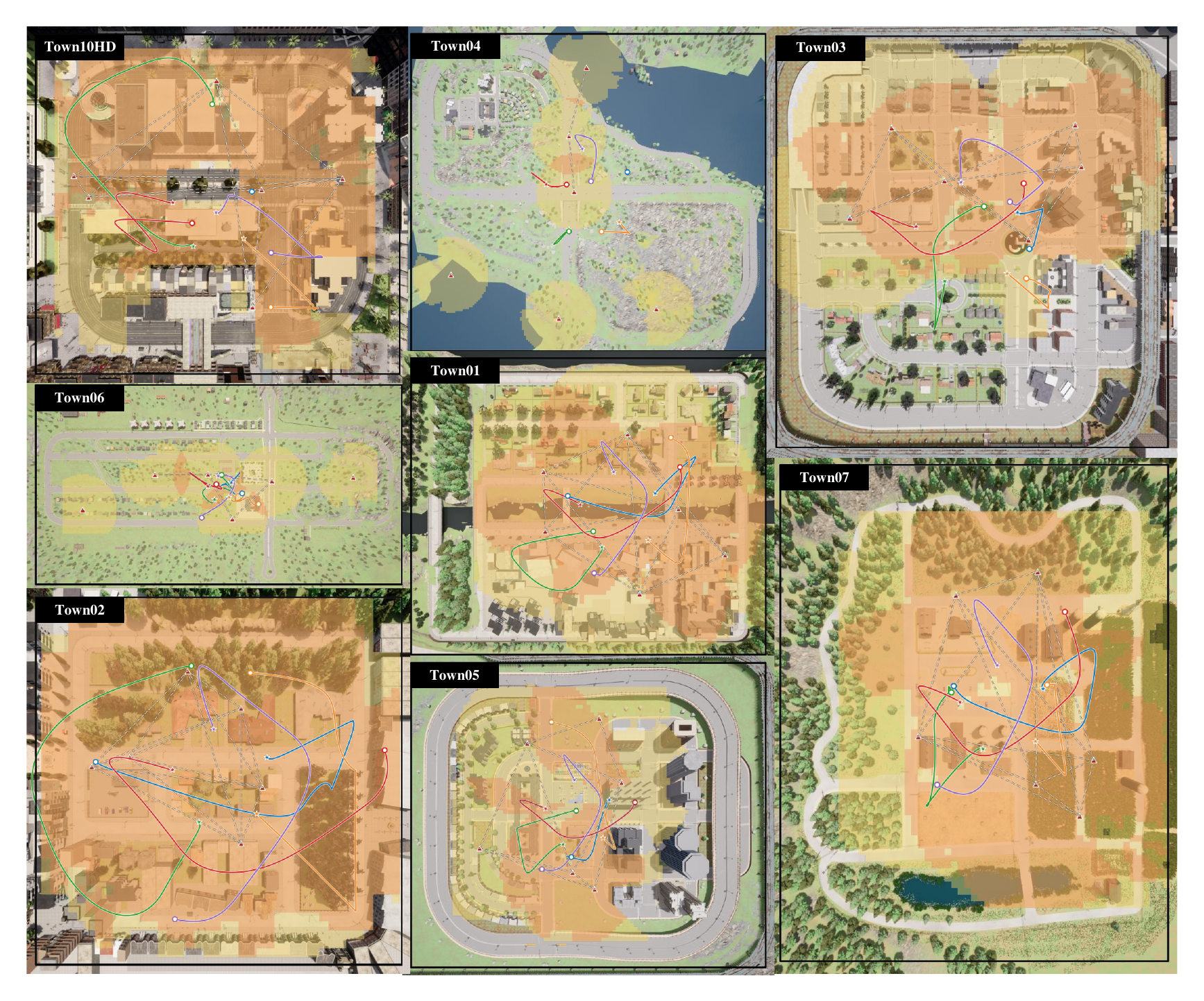}}
    \includegraphics[width=0.9\linewidth,height=\ht0]{fig/mapsInfo_pdf.pdf}
    \caption{Top-down overviews of TAO deployments on the eight maps under a budget of six cameras. Each panel contains five UAV trajectories and visualizes coverage voxels (yellow and orange), with orange indicating the robustly covered voxels. Although all cameras share the same viewing range, the maps differ in physical size; therefore, the panels are shown at different proportional scales.}
    \label{fig:maps_info}
\end{figure*}

\section{Deployment Visualization}
\label{sec:cross_map_generalization}

Figure~\ref{fig:maps_info} visualizes the TAO deployments on all eight CARLA
maps when $|C|=6$. The maps differ substantially in physical extent, road topology,
building density, and the spatial distribution of feasible camera locations.
Because the camera viewing range is fixed across all maps, the same camera
budget yields markedly different coverage fractions across environments.

\newpage

On expansive maps such as Town04 and Town06, Figure~\ref{fig:maps_info} shows
that TAO uses the limited camera budget to prioritize coverage while creating
overlap near the boundaries of neighboring camera FoVs whenever possible.
Such boundary overlap helps preserve the connectivity of the camera graph,
thereby increasing the likelihood of successful inter-camera handovers and
supporting longer continuous tracking durations.

\clearpage

\end{document}